\documentclass[10pt,twocolumn,letterpaper]{article}

\usepackage{ijcb}
\usepackage{times}
\usepackage{epsfig}
\usepackage{graphicx}
\usepackage{amsmath}
\usepackage{amssymb}
\usepackage{enumitem}

\makeatletter
\newcommand*\bigcdot{\mathpalette\bigcdot@{.5}}
\newcommand*\bigcdot@[2]{\mathbin{\vcenter{\hbox{\scalebox{#2}{$\m@th#1\bullet$}}}}}
\makeatother

\usepackage[pagebackref=true,breaklinks=true,letterpaper=true,colorlinks,bookmarks=false]{hyperref}

\ijcbfinalcopy 

\ifijcbfinal\fi
\begin{document}

\title{Anomaly Detection-Based Unknown Face Presentation Attack Detection}

\author{Yashasvi Baweja \hspace{6mm} Poojan Oza \hspace{6mm} Pramuditha Perera \hspace{6mm} Vishal M. Patel\\
Department of Electrical and Computer Engineering\\
Johns Hopkins University, 3400 N. Charles St., Baltimore, MD 21218, USA\\
{\tt\small $\lbrace$ybaweja1,poza2,pperera3,vpatel36$\rbrace$@jhu.edu}
}

\maketitle
\thispagestyle{empty}

\begin{abstract}
Anomaly detection-based spoof attack detection is a recent development in face Presentation Attack Detection (fPAD), where a spoof detector is learned using only non-attacked images of users. These detectors are of practical importance as they are shown to generalize well to new attack types. In this paper, we present a deep-learning solution for anomaly detection-based spoof attack detection where both classifier and feature representations are learned together end-to-end.  First, we introduce a pseudo-negative class during training in the absence of attacked images. The pseudo-negative class is modeled using a Gaussian distribution whose mean is calculated by a weighted running mean. Secondly, we use pairwise confusion loss to further regularize the training process. The proposed approach benefits from the representation learning power of the CNNs and learns better features for fPAD task as shown in our ablation study.  We perform extensive experiments on four publicly available datasets: Replay-Attack, Rose-Youtu, OULU-NPU and Spoof in Wild to show the effectiveness of the proposed approach over the previous methods.  Code is available at: \url{https://github.com/yashasvi97/IJCB2020_anomaly}
\end{abstract}

\section{Introduction}
\label{sec:intro}

With the ubiquitous use of mobile phones and laptops, security of digital devices have attracted considerable interest in the research community.  Recent advances in deep learning and facial recognition has prompted developers to use face and fingerprint-based authentication mechanisms in digital systems. Biometric-based authentication methods are 
vulnerable to carefully designed spoof attacks.  Face presentation attack is one such spoofing technique against which many face recognition modules fail considerably\footnote{www.wired.com/2016/08/hackers-trick-facial-recognition-logins-photos-facebook-thanks-zuck/}$^{,}$\footnote{www.wired.com/story/hackers-say-broke-face-id-security/} \cite{marcel2014handbook}. In a face presentation attack, attacker(s) tries to fool the biometric system by presenting a face picture (or a video) of  the enrolled user with the intention of surpassing authentication. To defend against such attacks, the device needs to learn how to distinguish between attacked images from the bonafide presentation images. This problem is commonly referred as the face Presentation Attack Detection (fPAD). 

\begin{figure}[t!]
	\centering
	\includegraphics[width=\linewidth]{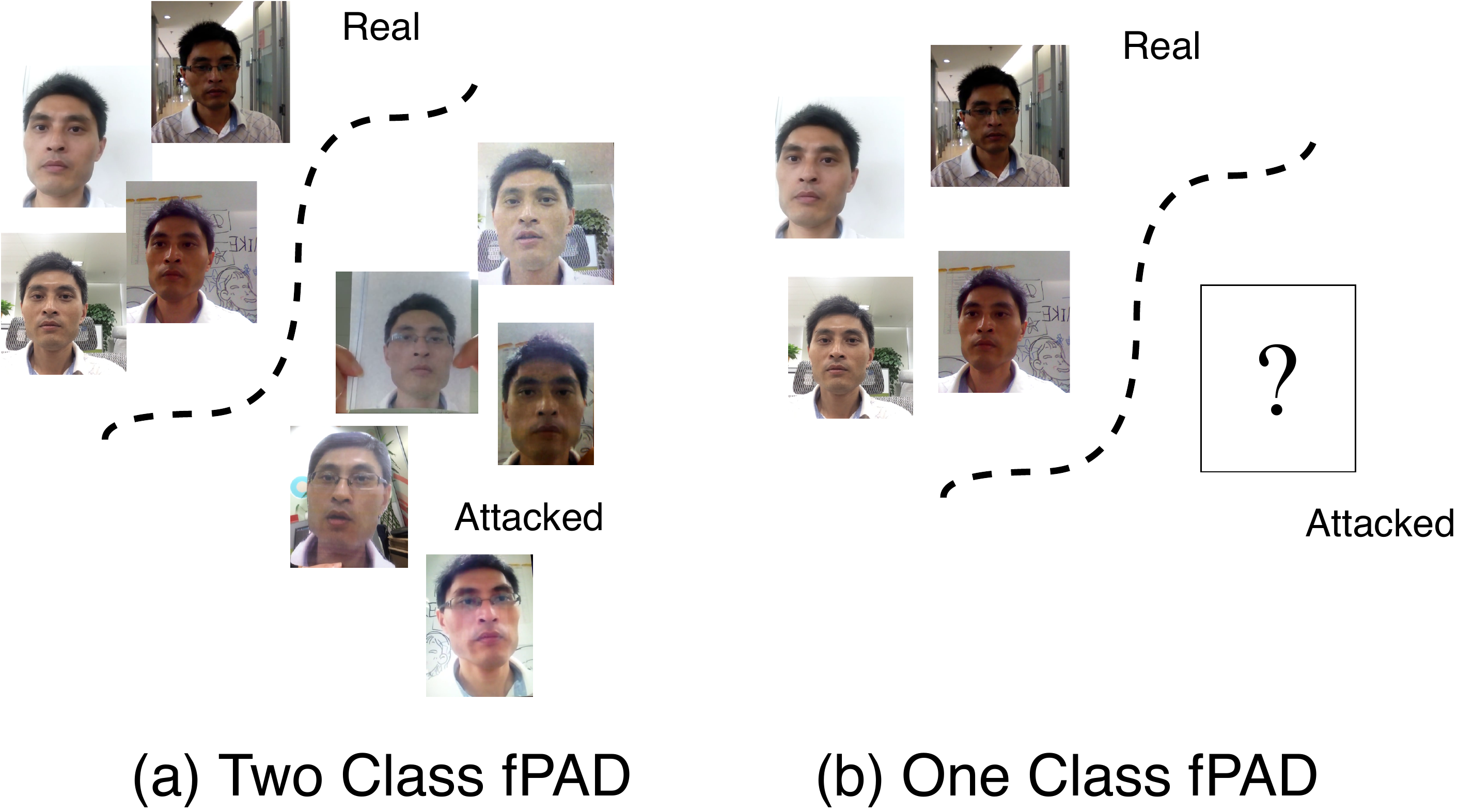}
	\caption{Different formulations  in fPAD. In (a) bonafide presentation images and attacked images of known attack types are available during training. A decision boundary is learned to distinguish between two types of images. (b) Only bonafide presentation images are available during training and a decision boundary containing all bonafide presentation data is learned.}
	\label{fig:motivation}
\end{figure}

Conventionally, fPAD has been formulated as a two class problem  in the literature \cite{george2019biometric, atoum2017face, liu2018learning,xiong2018unknown}. Specifically,  a binary classification model is trained using both attacked and bonafide presentation images (or other biometric cues) as shown in Figure~\ref{fig:motivation}(a). During inference, a query image is evaluated using the learned classification model. In this scenario, it is assumed that details of the attacking mechanism is known during training. In practice, attackers can potentially develop more sophisticated attacking mechanisms unknown to the device. Therefore, it is important that an fPAD mechanism is robust to attacks that are not seen during training. However, a study carried out by Arashaloo \etal \cite{arashloo2017anomaly} showed that binary classification-based fPAD methods do not generalize well to new attack types. Arashaloo \etal \cite{arashloo2017anomaly}  further showed that models with better generalization can be obtained by training a method only on bonafide presentation data (as done in anomaly detection) where known attack images are only used  for  model evaluation.  To this end,  \cite{arashloo2017anomaly} introduced a new branch of research in fPAD called \textit{anomaly detection-based fPAD}  where spoof detectors are learned only based on bonafide presentation images of the enrolled users, as illustrated in Figure~\ref{fig:motivation}(b).

Existing methods that address fPAD as an anomaly detection problem utilize off-the-shelf one class classification algorithms such as one class SVM (OC-SVM) \cite{scholkopf2001estimating}, Support Vector Data Descriptor (SVDD) \cite{tax2004support}, and one class Gaussian Mixture Models (OC-GMM) \cite{nikisins2018effectiveness}. These classifiers are learned on hand-crafted and/or deep features extracted from bonafide presentation images \cite{nikisins2018effectiveness, arashloo2017anomaly}. Fatemifar \etal \cite{fatemifar2019combining} utilized an ensemble of multiple one class classifiers to deal with this problem.  Recently, Fatemifar \etal \cite{fatemifar2019spoofing} proposed to include user id information to improve the fPAD performance. Note that these approaches either use hand-crafted  features or extract features from a pre-trained deep network.  These features are not learned directly from the training data.  Furthermore, none of these methods provide a standalone fPAD model that is end-to-end trainable.

In this paper, we propose an fPAD model that can be trained in an end-to-end fashion. Specifically, we utilize the representation learning power of a deep Convolutional Neural Networks (CNN) to learn better representation from given bonafide presentation image data and simultaneously learn classification boundary to enclose the learned representations.  Since only one class data (bonafide presentation images of the users) is available during training for the given problem, CNN training becomes a challenging task. To overcome this challenge, we propose a novel training strategy for CNNs by introducing pseudo-negative class samples in the feature space which  help the network learn a better decision boundary for fPAD.  In addition, we use a Pairwise Confusion (PC) loss \cite{dubey2018pairwise} to further regularize the fPAD network during training. 
 
This paper makes the following contributions:

\begin{enumerate}[noitemsep]
	\item An end-to-end  deep learning solution is presented for fPAD based on anomaly detection.
	\item We propose training deep CNNs with one class data for fPAD with the help of pseudo-negative sampling in the feature space.
	\item We perform extensive experiments on four publicly available datasets: Replay-Attack, Rose-Youtu, OULU-NPU and Spoof in Wild and show that the proposed approach is able to perform better compared to the existing fPAD methods.
\end{enumerate}

 
\section{Related Work} 
\label{sec:related}
\noindent \textbf{Traditional two class  based fPAD.} Traditional algorithms in fPAD  require a training dataset with labeled bonafide presentation and attacked images \cite{liu2018learning, boulkenafet2015face, atoum2017face}. Usually a binary classifier model is trained on this data and in many cases with the help of additional biometric cues \cite{atoum2017face, liu2018learning}, and finally the model predicts a test image as bonafide presentation or attacked based on the learned binary classification model. For example, Atoum \etal \cite{atoum2017face} introduced a novel end-to-end fPAD model which fuses scores from two deep CNNs -- one from depth-based and another from image patch-based. In some of the recent methods, Liu \etal in \cite{liu2018learning} trained a binary classifier for fPAD using other modalities, like rPPG signal, depth map, optical flow \etc.   George \etal \cite{george2019biometric} used a multi channel information (like depth, Infra-Red, \etc) across same feature extraction network and classify the image as bonafide presentation or attacked based on the concatenated output feature. 

 \begin{figure}[!t]
 	\centering
 	\includegraphics[width=\columnwidth]{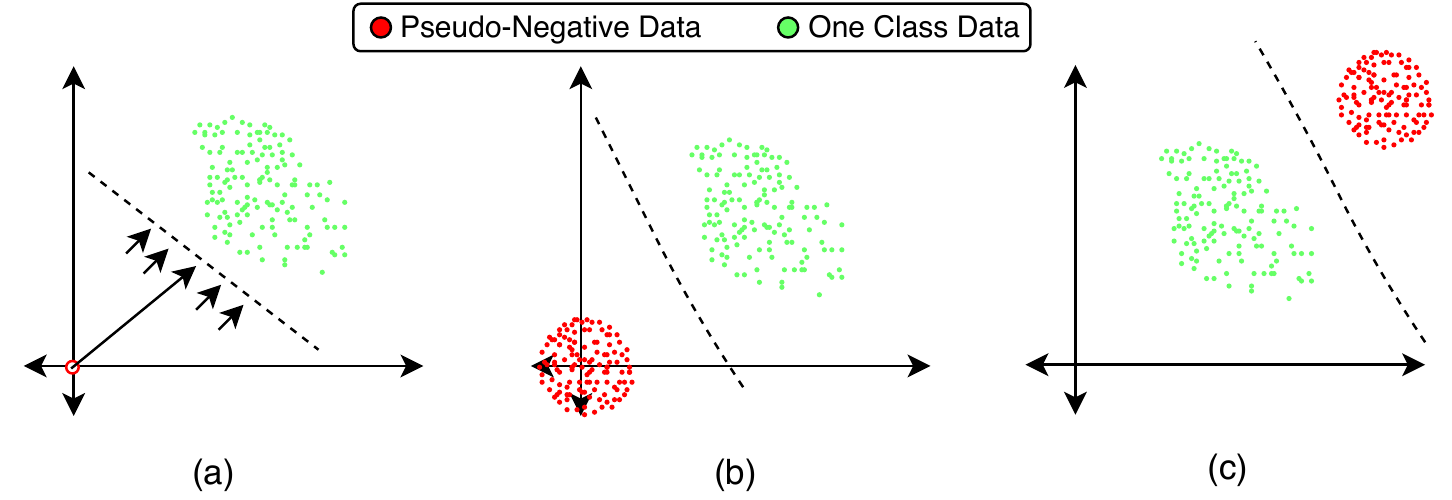}
 	\caption{Comparison of one class classification methods. (a) One Class SVM \cite{scholkopf2001estimating}. (b) OC-CNN \cite{oza2018one}. (c) Proposed method.}
 	\label{fig:ocnn_methods}
 \end{figure}

Several other studies use temporal information of videos to detect whether a video is attacked or not. Liu \etal \cite{liu20163d} use rPPG signal to differentiate between bonafide presentation and 3D mask attacked videos. Liu \etal \cite{liu2018remote} further extended the approach and used Near Infrared (NIR) to work with remote photoplethysmography correspondence features for detecting 3D mask attacked videos. Several works have used  texture-based analysis for detecting attacked images. Boulkenafet \etal \cite{boulkenafet2015face}  deploy Local Binary Patterns (LBP) \cite{ahonen2006face} features calculated for all RGB dimensions and use the concatenated LBP to classify it as bonafide presentation or attacked image.  Boulkenafet \etal improved this method later \cite{boulkenafet2016face} using SIFT \cite{lowe2004distinctive} and SURF \cite{bay2006surf} features in both RGB and YCbCr space. In, \cite{galbally2014face}  authors explored the possibility of using Image Quality Measure (IQM) features to perform fPAD.
\begin{figure*}
	\centering
	\includegraphics[width=\textwidth]{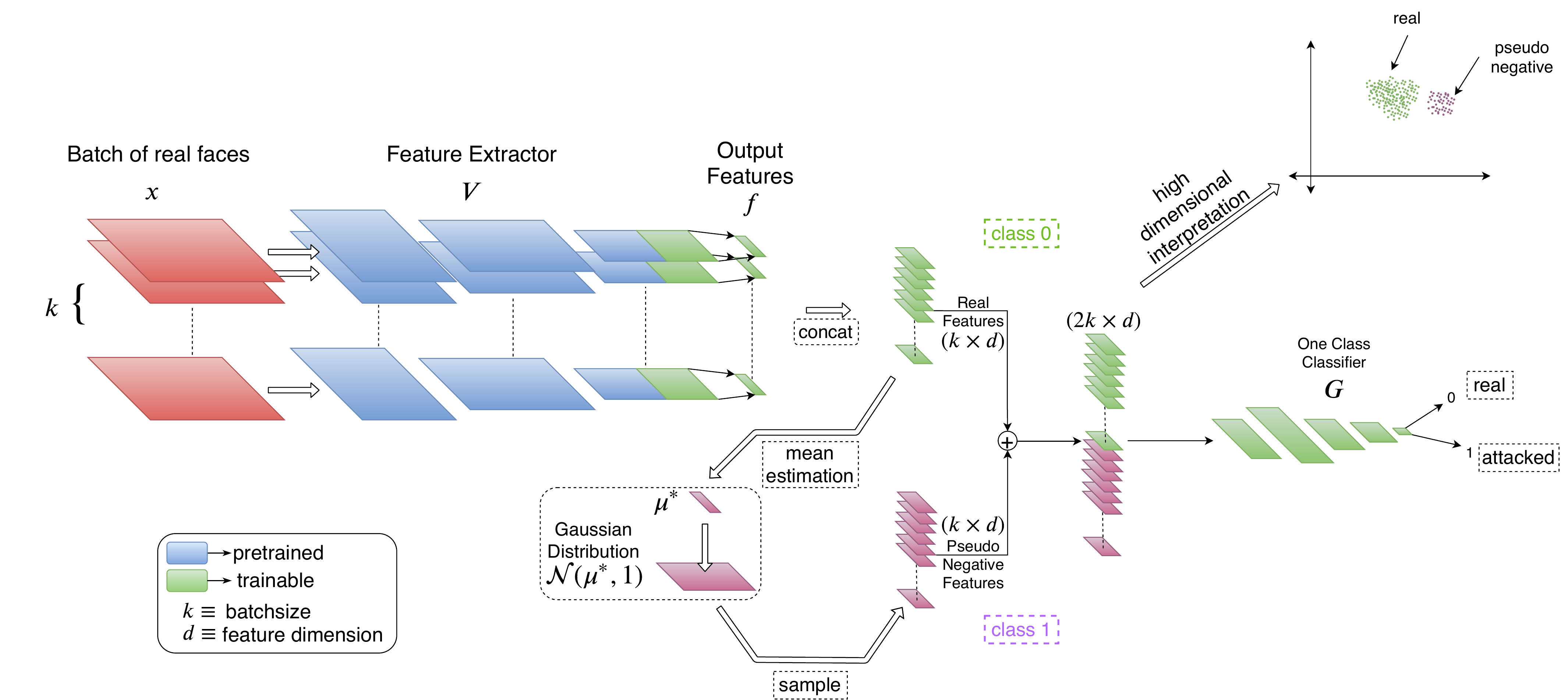}
	\vskip -2.0pt
	\caption{Training algorithm of the proposed method (best viewed in color). A batch of bonafide presentation images (red) is fed into the face extractor $V$, which in turn produces a batch of bonafide presentation features (green) of dimension ($k \times d$).  A pseudo-negative feature batch (purple)  $k \times d$   is  sampled from a Gaussian distribution whose mean is calculated considering a running mean of bonafide presentation features. The concatenated batch of bonafide presentation and pseudo-negative features is fed into the classifier. The output of the classifier $G$ produces the posterior probability of bonafide presentation and attacked classes.}
	\label{fig:proposed}
\end{figure*}

\noindent \textbf{Anomaly Detection-based fPAD.} Arashloo \etal \cite{arashloo2017anomaly} argued that such two-class approaches can be biased towards attacked training images and might limit the generalization ability of the model to novel attacks. Hence, \cite{arashloo2017anomaly}  provided a novel formulation for fPAD based on anomaly detection where models are learned using only bonafide presentation images. Following this formulation, Arashloo \etal \cite{arashloo2017anomaly} proposed a one class SVM \cite{scholkopf2001estimating} based method for fPAD. Nikisins \etal \cite{nikisins2018effectiveness} proposed another approach by modeling Image Quality Measure (IQM) features \cite{galbally2013image} extracted from the bonafide presentation images with one class GMM and showed superiority over the one class SVM classifier. Fatemifar \etal \cite{fatemifar2019spoofing} showed that when client ID information is available, it can be leveraged to further boost the fPAD performance. Specifically, they train client-specific PAD models that are based on one class SVM, one class GMM and Mahalanobis distance. Fatemifar \etal \cite{fatemifar2019combining} explored the use of features from specific face regions such as eyes, mouth, nose etc. which might be helpful in improving the fPAD performance. Another approach by Fatemifar \etal \cite{fatemifar2019combining} utilized ensembles of one class classifiers. All of these methods use hand-crafted features or features from a pre-trained deep network and learn an off-the-shelf one class classifier. None of these methods have utilized deep networks in the training process to leverage their representation learning power. 

\noindent \textbf{One Class Methods} Oza \etal \cite{oza2018one} introduce new one class algorithm where a pseudo negative class centered at origin is used to train the one class classifier. Authors extended the same one class approach for active authentication in \cite{oza2019active}. Perera \etal \cite{perera2018dual} also deploy the one class classification method for active authentication. Authors in \cite{perera2019learning} learn sophisticated deep features focused for one class classification. Open set algorithms also aim to detect anomalies or out of distribution data: Perera \etal \cite{perera2019deep} introduce membership loss for novelty classification. Authors in \cite{oza2020novelty, oza2020patch} use data distribution shift and patch level information respectively for multiple class novelty detection. Authors in \cite{perera2020generative} learn discriminative deep features for open set recognition. Zhang \etal \cite{zhang2016sparse} induce a sparsity constraint on features for open set classification. Further, Shao \etal \cite{rui2020open} defend against open set adversarial attacks.

\section{Proposed Method}
\label{sec:proposed}
For one class problems, the absence of negative class data makes it difficult to train a deep network in an end-to-end fashion. Hence, most work follow a methodology where deep features are extracted using a pre-trained network and a one class classifier such as one class SVM (OC-SVM) is trained for classification.  To enable end-to-end learning for deep networks in one class setting, Oza and Patel \cite{oza2018one} proposed a method named one class convolutional neural network (OC-CNN) which attempts to mimic the idea behind OC-SVM where a separating hyperplane is learned to separate one class features from the origin. Specifically, they use samples from a Gaussian distribution centered at the origin with a small standard deviation as the pseudo-negative class. Inspired from this work, we utilize similar framework for training the deep networks. However, the problem setting considered in \cite{oza2018one} considers anomaly data that are visually distinct as compared to the training class data we have in fPAD. Hence, the concept of separating training data in the feature space from a zero centered Gaussian might work well for their problem setting. Whereas, in the case of fPAD the attacked samples during testing are very similar to the non-attacked samples used during training. As a result there exists only a subtle visual difference between the two classes (bonafide presentation and attacked images of the same person). Hence, images from both classes can be in a very close proximity to each other in the feature space. To deal with this issue, we propose an adaptive mean estimation strategy to generate pseudo-negative data for training. The goal of this adaptive strategy is to create the pseudo-negative Gaussian distribution such that it lies in the close proximity to the features from the bonafide presentation class data. The differences among OC-SVM, OC-CNN and the proposed method are illustrated in Figure~\ref{fig:ocnn_methods}. In the following sub-sections we discuss the proposed method in detail. The proposed method consists of two main parts, a feature extractor network ($V$) and a one class classifier ($G$).

\subsection{Feature Extractor}
\label{subsec:feat_ext}
Let $\{x_{i}\}_{i=1}^{N}$ be a set of $N$ training images consisting of bonafide presentation (i.e. non-attacked) images. These images are passed through a face recognition network $V$ which produces a set of $d$-dimensional features $\{f_{i}\}_{i=1}^{N}$ such that $f_i = V(x_i)  \in \mathbb{R}^{d}$. These features are then fed into the one class classifier, $G$, as described in the following Section.

\subsection{One Class Classifier}
\label{subsec:oc_nn}


As discussed earlier, we utilize an adaptive strategy to estimate the mean of pseudo-negative Gaussian distribution. The adaptive mean estimation makes sure that samples from pseudo-negative Gaussian distribution lies in the proximity of bonafide presentation class features. More formally, we define a pseudo-negative distribution,  $\mathcal{N}(\mu^{*}, \sigma)$ where  $\mathcal{N}$ is a Gaussian distribution whose mean is $\mu^{*}$ and standard deviation is $\sigma$.  We consider displacement of sample feature mean of bonafide presentation data across two iterations when defining $\mu^{*}$. More precisely, let us consider a batch ($B$) that contains features extracted from the non-attacked images using network $V$, denoted as $f^B = \lbrace f_i \rbrace, \forall i \in B$. Let $\mu_{new}$ be the mean of feature vectors $f^{B}$ and $\mu_{old}$ be the mean of features of the previous batch. Both $\mu_{new}$ and $\mu_{old}$ are $d$-dimensional. For training a one class classifier $G$ a batch $B'$ of attacked features $f^{B'} = \lbrace f_j \rbrace, \forall j \in B'$ are sampled from the pseudo-negative class, whose center $\mu^{*}$ is calculated as 
\setlength{\belowdisplayskip}{0pt} \setlength{\belowdisplayshortskip}{0pt}
\setlength{\abovedisplayskip}{0pt} \setlength{\abovedisplayshortskip}{0pt}
\begin{equation}
	\mu^{*} = \alpha \mu_{old} + (1 - \alpha) \mu_{new},
\end{equation}
where $\alpha$ is a hyper-parameter controlling the influence of old mean in the estimation of $\mu^{*}$ and  $| B' | = | B | = k$. During the  first iteration $\mu_{new}$ is used to calculate $\mu^{*}$. The data $f = \left(f^{B} , f^{B'}\right)$ is concatenated across the batch dimension and fed into the classifier $G$. The corresponding label vector, is of size $2k \times 1$, where the first $k$ elements are all zeros followed by all ones denoting bonafide presentation and attacked class, respectively. This  process is illustrated in Fig.~\ref{fig:proposed}. The classifier $G$ uses these features and produces probability vector $p$ of size $2k \times 1$. The network can be trained using the cross-entropy loss defined as follows
\begin{equation}
\ell_{ce} = - \sum_{i=1}^{2k} \lbrace y_i \log(p_i) + (1 - y_i) \log(1 - p_i) \rbrace,
\end{equation}
where, $y_i$ denotes the label (0 or 1) of the $i^{th}$ input to the classifier $G$ and $p_i$ denotes the probability of the $i^{th}$ input being from the pseudo-negative class. Similarly, $1-p_i$ denotes the probability of the $i^{th}$ input to the classifier $G$ being extracted from the non-attacked class data. Our classifier $G$ consists of 3 fully-connected layers of 8192, 1000, 500 neurons, respectively. The dimension of the input layer is the same as the output layer of $V$.  The final output has two neurons, one for the probability corresponding to the bonafide presentation class and other for the attacked class.

\begin{figure*}[t!]
	\centering
	\includegraphics[width=\linewidth]{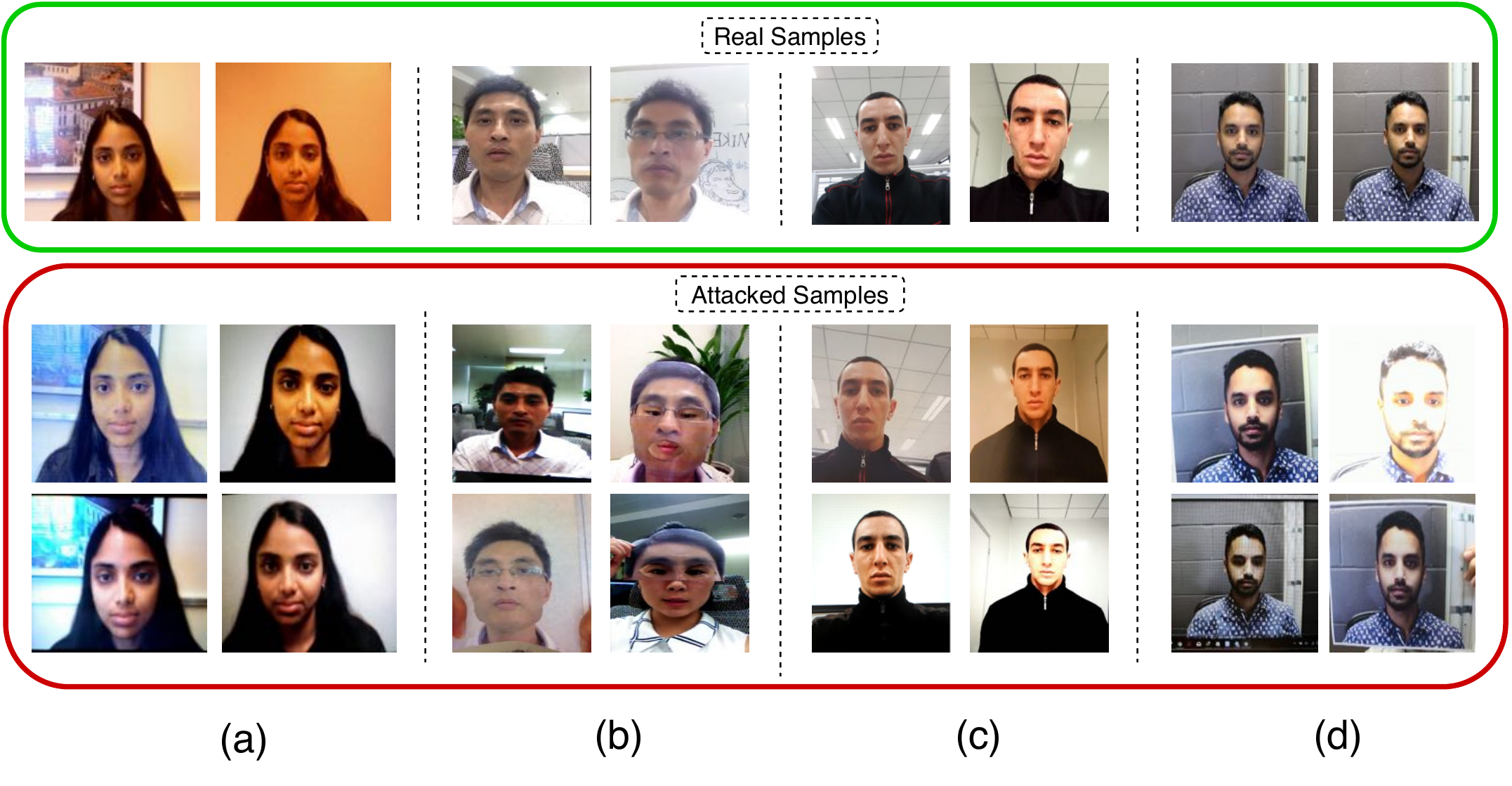}
	\vskip -8.0pt
	\caption{Sample images from each dataset. (a) Replay Attack \cite{replay_attack}, (b) Rose-Youtu \cite{rose_youtu} (c) Oulu-NPU \cite{oulu_npu} and (d)Spoof in Wild \cite{liu2018learning}. Green part represents the bonafide presentation samples and red corresponds to the attacked samples of same identity.}
	\label{fig:datasets}
\end{figure*}

As in most cases the network $V$ is pre-trained on a face recognition dataset and in this case it's VGGFace \cite{parkhi2015deep}. Although pre-trained weights serve as a good initialization for training the initial layers of $V$, the high level features (which will be used as input to $G$) are not suitable for fPAD. The final layer features of a pre-trained network are more suitable for face recognition.  Furthermore, feature representations are usually clustered by identities.  Eliminating the identity information from the features helps in improving the fPAD performance as we are only interested in spoof detection irrespective of the identity information. To this end, we utilize the following Pairwise Confusion (PC) loss \cite{dubey2018pairwise}
\begin{equation}
\ell_{pc} = \sum_{i} \sum_{j \neq i} \| f_i - f_j \|_{2}^{2},
\end{equation}
where $f_i$ is the feature vector corresponding to image $x_i$. This loss induces a Euclidean confusion in the pre-trained feature space and tries to remove the identity information from it. Disrupting the feature space in this manner results in better feature representation for fPAD. Note that, this loss is only calculated using the bonafide presentation features and not from the pesudo-negative Gaussian samples. 

The overall loss function used to train the network is a linear combintion of both loss functions and is defined as follows  
\begin{equation}
\label{final_loss}
\ell_{final} = \lambda_1 \ell_{pc} + \lambda_2 \ell_{ce},
\end{equation}
where $\lambda_1$ and $\lambda_2$ are two constants.  This loss function is used to train the spoof detection network in an \textit{end-to-end} manner.

\section{Datasets and Evaluation Protocols}
\label{sec:datasets}
We demonstrate the effectiveness of the proposed model for fPAD on four datasets: Replay Attack \cite{replay_attack}, Rose-youtu \cite{rose_youtu}, OULU-NPU \cite{oulu_npu} and Spoof in Wild \cite{liu2018learning}. Sample images from each dataset are shown in Figure \ref{fig:datasets}.  

\subsection{Replay Attack}
The Replay Attack dataset \cite{replay_attack} is an fPAD dataset of 1300 videos consisting of photo and replay attacks. It contains 50 identities and four sets namely, train, development, test and enroll. The train, development and test sets contain both bonafide and attacked presentation data of 15, 15 and 20 identities, respectively. We extract 30-40 frames per video with a gap of atleast 10 frames in between to carry out our experiments. HOG-based face detection \cite{dalal2005histograms} algorithm is then used to find the face.  We train our model and the baselines on only bonafide presentation data from the training set and test on the bonafide presentation and all attacked data from the test set.

\subsection{Rose-Youtu}
The Rose-youtu dataset \cite{rose_youtu} is an fPAD database which covers a large variety of illumination conditions and attack types. There are in total 3350 frontal face videos of 20 subjects captured using 5 mobile phones: Hasee smart-phone, Huawei smart-phone, iPad 4, iPhone 5s, and ZTE smart-phone.  On average each subject has around 25-50 bonafide presentation videos. There are three types of attacks in the dataset: print attack, replay video attack and mask attack. Every 15th frame from the video under consideration is sampled and then corresponding face is extracted using the HOG-based face detector \cite{dalal2005histograms}. We use 10 train identities' bonafide presentation data for training the baselines and the proposed method. Testing is done on the data from the remaining 10 identities using all attacks.

\subsection{OULU-NPU}
The OULU-NPU dataset \cite{oulu_npu} consists of mobile attacked videos. There are in total 4950 videos, 55 identities recorded using six mobile devices in different sessions. There are mainly two types of attacks in the dataset -- print and replay attack. The print attacks are done using two types of printers and video replays are done using two display devices. We consider all attacks in the test set. For training, only bonafide presentation data corresponding to the identities in training set are used. For extracting faces from videos, we sample every 10th frame from each video and then use the face locations given in the dataset. There are 20 and 15 different identities in the training set and the test set, respectively. 

\subsection{Spoof in Wild}
The Spoof in Wild dataset \cite{liu2018learning} is a collection of 4478 videos distributed among 165 identities. All videos are captured in minimum 1080p resolution. Each subject has 8 bonafide presentation and around 20 attacked videos. There are two types of attacks in the dataset -- print attack based on glossy and matt paper, and replay attack created using 4 different devices. All attacks are considered during testing. Faces are extracted from each video by sampling at every 10th frame. We use the face locations provided in the dataset to extract faces. This data is then used to carry out experiments. We use only bonafide presentation data from the training set for training the baselines and the proposed model. Data from 90 subjects are used during training and the remaining data from 75 different users are used during teting.

We evaluate the performance of anomaly detection-based fPAD methods on two protocols. The details of these protocols are described as follows:\\
$\bullet$\textbf{ Protocol 1:} In this protocol models are trained using the identities in the training set with only bonafide presentation image data and evaluated on the identities of the test set having both bonafide presentation and attacked images. Both sets contain non-overlapping identities.\\
$\bullet$ \textbf{Protocol 2:} For this protocol, enrollment split is used for model training and evaluation \cite{fatemifar2019spoofing}. fPAD models are trained using a training set and evaluated on the test set. Both training and test set are derived from the enrollment split of the datasets and both have bonafide presentation and attacked images of same identities. Unlike \cite{fatemifar2019combining}, we do not utilize any identity label information and hence for fair comparison consider only class-independent models of \cite{fatemifar2019combining}.

\begin{table*}[!t]
	\centering
\begin{tabular}{|c|c|c|c|c|c|c|}
	\hline
	\textbf{Dataset} & \textbf{OC-SVM} & \textbf{SVDD} & \textbf{MD} & \textbf{OC-GMM} & \textbf{OC-CNN} & \begin{tabular}[c]{@{}c@{}}\textbf{Proposed}\\ (APCER, BPCER)\end{tabular} \\\hline \hline
	Replay Attack \cite{replay_attack} & 31.142 & 32.961 & 31.747 & 30.096 & 35.985 & \begin{tabular}[c]{@{}c@{}}\textbf{20.739}\\ (25.047, 16.539)\end{tabular} \\ \hline
	Rose Youtu \cite{rose_youtu} & 47.165 & 46.532 & 38.993 & 46.771 & 35.808 & \begin{tabular}[c]{@{}c@{}}\textbf{31.623}\\ (35.573, 27.673)\end{tabular} \\ \hline
	Oulu NPU \cite{oulu_npu} & 47.561 & 47.517 & 45.185 & 46.957 & 45.799 & \begin{tabular}[c]{@{}c@{}}\textbf{30.242}\\ (38.632, 21.852)\end{tabular} \\ \hline
	Spoof In Wild \cite{liu2018learning} & 47.165 & 47.051 & 42.970 & 44.250 & 36.994 & \begin{tabular}[c]{@{}c@{}}\textbf{23.335}\\ (23.482, 23.187)\end{tabular} \\ \hline
\end{tabular}
	\caption{Comparison of the proposed algorithm with baseline one class classifiers for Protocol 1. The values in bold represent the lowest ACER(\%) with APCER(\%) and BPCER(\%) in brackets.}
	\label{tab:baseline}
\end{table*}

\section{Implementation Details}
In all datasets we only use the bonafide presentation data from the training set without using any identity label information for training the model. The test data consists of both bonafide and attack presentation data. For datasets without face locations, we use HOG-based face detection mechanism \cite{dalal2005histograms}. The values for $\lambda_1$ and $\lambda_2$ are set equal to 3 and 1, respectively. The value of $\alpha$ is set equal to 0.8. The standard deviation of the pseudo-negative Gaussian is set as 1. VGGFace \cite{parkhi2015deep} is used as the base feature extraction module. The layer just after the convolution layers, that is `fc6'(4096 dimension) of VGGFace is used to extract features. Thus output of the `fc6' layer is used as the input to the classifier $G$.  The last two convolution layers, `fc6' of $V$ along with all the parameters of $G$ are trained using the loss equation \eqref{final_loss}. The model is trained for 100 epochs with 1e-4 learning rate, batch size of 80 and results are reported in Section~\ref{sec:results}.

\section{Results and Analysis}
\label{sec:results}
In this section, we discuss results on the aforementioned datasets and present an in depth analysis of the proposed method. We use Average Classification Error Rate (ACER) as the metric to compare the performance of different methods.  ACER is the average of the Attack Presentation Classification Error Rate (APCER) and Bonafide Presentation Classification Error Rate (BPCER) at a particular threshold \cite{ramachandra2017presentation}. Since APCER and BPCER are threshold-based values, ACER is also dependent on a threshold $\tau$ and is defined as follows
\[ ACER = \frac{APCER + BPCER}{2} \% .\]
In our comparisons we aim to find the threshold for which the average of APCER and BPCER is least. ACER is calculated for each identity in the test set and the average of ACER is reported in Table \ref{tab:baseline}.

\subsection{Comparison with Baseline One Class classifiers}
For our baseline comparisons we consider the following five methods:

\begin{enumerate}
	\item \textbf{OC-SVM} - One class SVM (OC-SVM) \cite{scholkopf2001estimating} maps the one class data into a feature space and tries to learn a hyperplane between the training data and origin. Implementation from sklearn library \cite{scikit-learn} is used with default parameters and $\nu$=0.1. Signed distance from the hyperplane is used as an anomaly score.
	\item \textbf{SVDD} - Support Vector Data Description (SVDD) \cite{tax2004support} is an extension to the one class SVM where it tries to learn a hypershpere around the training data as opposed to a hyperplane separating the origin. Optimization is done to have minimum spherical radius that can contain all data. Default implementation of LibSVM library \cite{libsvm} is used with parameter $\nu$=0.1. Distance from the center is used as the anomaly score.
	\item \textbf{MD} - Mahalanobis Distance (MD) assumes data is coming from a Gaussian distribution with single mode and computes the distance based on the parameters of the Gaussian model. The bonafide presentation data is used to estimate the parameters of Gaussian model. The distance calculated based on the estimated mean and variance of Gaussian model is used as the anomaly score. Higher distance increases the likelihood of a test image being an attacked image. MD is implemented using the Sklearn library.
	\item \textbf{OC-GMM} - A Gaussian Mixture Model (GMM) is learned using bonafide presentation class data. Parameters of the GMM are estimated using the Sklearn library. For consistency across all datasets, the number of components are set equal to 50. Log likelihood of a test sample belonging to GMM is used as the anomaly score for calculating ACER.
	\item \textbf{OC-CNN} - One Class CNN \cite{oza2018one} trains a classifier using one class data and pseudo-negative Gaussian distribution centered at the origin. The probability that a test sample is a bonafide presentation is used as the anomaly score.
\end{enumerate}
For the proposed algorithm, we use the probability for class 0, \ie, being a bonafide presentation as the anomaly score to calculate ACER.

\subsection{Experiments using Protocol 1}
As can be seen from Table~\ref{tab:baseline} the proposed algorithm (last column) outperforms all the other methods. Among the baseline methods, one class GMM performs the best as it helps capturing the multi-modal nature of the distribution of bonafide presentation features. However, the proposed approach outperforms one class GMM significantly, providing overall improvement of $\sim$6\% across all datasets. This performance improvement can be attributed to the novel training strategy of separating the bonafide presentation features from Gaussian sampled features, which are in close proximity and helps in learning a good decision boundary. Furthermore, training a deep neural network using the given data for spoof detection helps in learning more rich features rather than using hand-crafted or pre-trained features.

\subsection{Experiments using Protocol 2}
\begin{figure}[!b]
	\includegraphics[width=\linewidth]{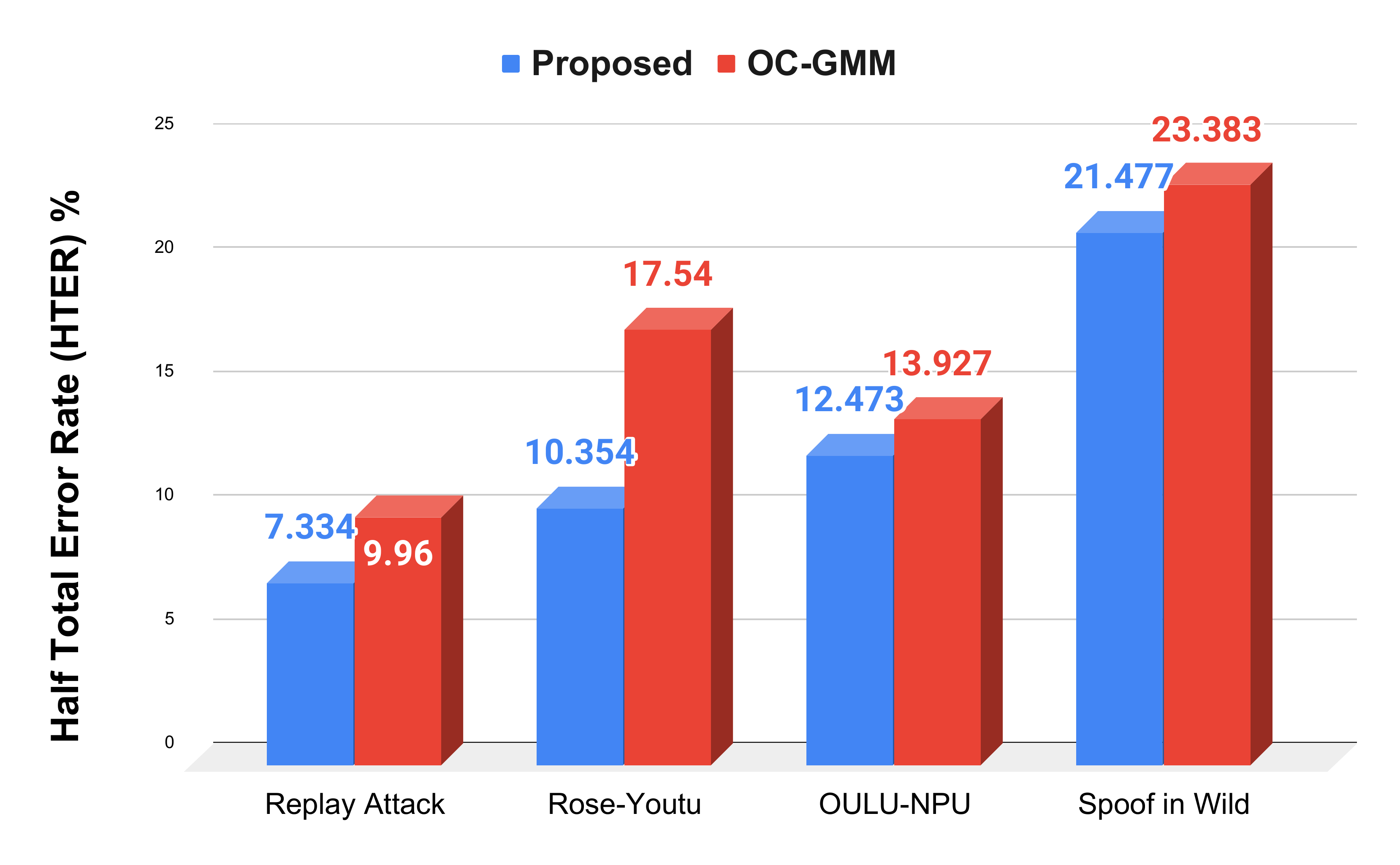}
	\vskip -8.0pt
	\caption{Comparison of the best baseline (OC-GMM) and the proposed method for Protocol 2 in terms of ACER.}
	\label{fig:diff_eval}
\end{figure}

The results of experiments on protocol 2 are summarized in Fig.~\ref{fig:diff_eval}. Specifically, for the Replay Attack dataset \cite{replay_attack} the best identity independent one class classifier is one class GMM, which achieves ACER of 9.96\% \cite{fatemifar2019combining}. Whereas the proposed method achieves ACER of 7.334 \% providing $\sim$2\% improvement. For the Rose-Youtu dataset, one class GMM achieves ACER of 17.54\% \cite{fatemifar2019combining}, whereas the proposed algorithm achieves ACER of 10.354\%, resulting in an improvement of $\sim$7\%. Additionally, we performed protocol 2 experiments on OULU-NPU and Spoof in Wild and the best performing classifier OC-GMM achieved ACER of 13.927\% and 23.383\%, respectively. In comparison, the proposed approach achieved ACER of 12.473\% and 21.557\% for OULU-NPU and Spoof in Wild, providing $\sim$1\% and $\sim$2\% improvements respectively.

The results in Table ~\ref{tab:baseline} and Figure \ref{fig:diff_eval} show that the proposed approach outperforms other one class classifiers in both Protocol 1 and Protocol 2.

\subsection{Comparison with OC-CNN}
The method proposed in \cite{oza2018one}, introduced a pseudo-negative class to train the one class classifier. The method works well for their problem setup of anomaly detection, where training and test data have categorical and visual differences. But considering the fPAD-based anomaly deteting problem formulation, the face images at training and test time have very subtle visual differences. In some cases, these images may belong to the same identity and therefore it will not be optimal to separate bonafide presentation features from $\mathcal{N}(0, 1)$ noise. From  the results in Table~\ref{tab:baseline}, when the model is trained with a Gaussian centered at 0, ACER value is 35.985\% for the Replay Attack Dataset. Instead when trained with a Gaussian whose center is chosen by the running mean, i.e., the proposed method, we see a significant improvement performance, achieving ACER of 20.739 \%. Therefore, modeling the attacked class in the proximity to the bonafide presentation features helps in building a more sophisticated one class classifier for fPAD.

\begin{figure}[!t]
	\includegraphics[width=\columnwidth]{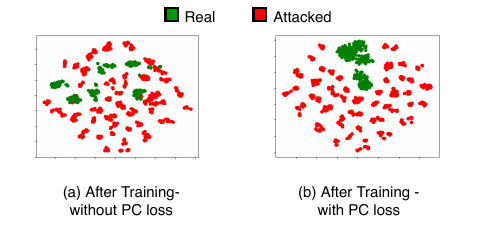}
	\caption{t-SNE plots for data corresponding to identities 1, 2, and 4 of the Replay-Attack dataset combined. The points in green represent the bonafide presentation features, and their attacked versions are marked in red. Figure (a) shows the t-SNE plot of the feature space when extracted using a trained network without the PC Loss. It can be seen that all the bonafide presentation features are cluttered in the feature space and are closer to their attacked counterparts. Figure (b) on the other hand shows the same feature space, but now the bonafide presentation features are all packed into a single cluster which are more suited for one class seting.}
	\label{fig:tsne_ocnn}
\end{figure}

\subsection{Ablation study}

\textbf{Effect of Pairwise Confusion loss}: In this experiment we study the effect of pair-wise confusion loss \cite{dubey2018pairwise} on the performance of the proposed model. We show that when the model is trained without the pair-wise loss, the performance is lower as compared to when PC loss is used. Here, we consider Replay-Attack dataset for the experiments. When the proposed model is trained with the PC loss, the performance is 20.739\% ACER. The t-SNE feature visualization for this experiment is also provided in Fig.~\ref{fig:tsne_ocnn}(b). On the other hand, when we train the proposed method without the pair-wise confusion loss, the ACER increases to 32.860\%. The corresponding t-SNE feature visualization for this experiment is given in Fig.~\ref{fig:tsne_ocnn}(a). When we compare the t-SNE visualization of the features provided in Fig.~\ref{fig:tsne_ocnn}, we can see that the best the model trained with pair-wise confusion loss achieves the feature embedding most suited for the anomaly detection-based fPAD model and thereby resulting in the improved performance among all other baselines.

\section{Conclusion}
\label{sec:conclusion}
In this paper we tackled the problem of face presentation attack detection based on anomaly detection. We point out the shortcomings of the existing work that address this issue and propose a novel method to alleviate these shortcomings. More precisely, we proposed a new end-to-end trainable, face presentation attack detection model based on deep convolutional neural network that can be trained with data from only one class. The proposed method utilizes a novel training procedure for one class neural network with the help of pseudo-negative sampling in the feature space. To the best of our knowledge, this is the first approach to include deep networks in the learning process and leverage their power of representation learning in an end-to-end manner for anomaly-based face presentation attack detection task. Extensive experiments on four publicly available datasets show that the proposed approach outperforms all existing approaches for the task. Furthermore, an ablation study was provided to show the significance of individual components of the proposed method.


\section*{Acknowledgment}
This work was supported by the NSF grant 1801435.

{\small
\bibliographystyle{ieee}
\bibliography{submission_example}
}

\end{document}